\documentclass[sigconf]{acmart}

\AtBeginDocument{%
  \providecommand\BibTeX{{%
    \normalfont B\kern-0.5em{\scshape i\kern-0.25em b}\kern-0.8em\TeX}}}

\copyrightyear{2021}
\acmYear{2021}
\setcopyright{acmlicensed}
\acmConference[HT '21]{Proceedings of the 32nd ACM Conference on Hypertext and Social Media}{August 30-September 2, 2021}{Virtual Event, Ireland}
\acmBooktitle{Proceedings of the 32nd ACM Conference on Hypertext and Social Media (HT '21), August 30-September 2, 2021, Virtual Event, Ireland}
\acmPrice{15.00}
\acmDOI{10.1145/3465336.3475102}
\acmISBN{978-1-4503-8551-0/21/08}

\begin{document}

\fancyhead{}

\title{Cross-lingual Capsule Network for Hate Speech Detection in Social Media}

\author{Aiqi Jiang}
\affiliation{%
  \institution{Queen Mary University of London}
  \streetaddress{Mild End Road}
  \city{London}
  \country{UK}}
\email{a.jiang@qmul.ac.uk}

\author{Arkaitz Zubiaga}
\affiliation{%
  \institution{Queen Mary University of London}
  \streetaddress{Mild End Road}
  \city{London}
  \country{UK}}
\email{a.zubiaga@qmul.ac.uk}

\renewcommand{\shortauthors}{A. Jiang and A. Zubiaga}

\begin{abstract}
Most hate speech detection research focuses on a single language, generally English, which limits their generalisability to other languages. In this paper we investigate the cross-lingual hate speech detection task, tackling the problem by adapting the hate speech resources from one language to another. We propose a cross-lingual capsule network learning model coupled with extra domain-specific lexical semantics for hate speech (CCNL-Ex). Our model achieves state-of-the-art performance on benchmark datasets from AMI@Ev-alita2018 and AMI@Ibereval2018 involving three languages: English, Spanish and Italian, outperforming state-of-the-art baselines on all six language pairs.

\end{abstract}



\maketitle

\section{Introduction}

Anonymity and lack of moderation provide benefits for social media, alongside negative effects such as production of harmful and hateful contents \citep{schmidt2017survey,fortuna2018survey,yin2021towards}. The necessity to tackle hate speech has attracted the attention of the scientific community, industry and government to come up with automated solutions. Most existing research focuses on English hate speech, due to its advantageous position over other languages in terms of available resources. This in turn leads to a dearth of research in other languages \cite{fortuna2018survey}. Despite the recent trend of increasingly investigating hate speech detection in other languages \citep{vidgen2020directions}, most of the work is limited to a single language. As far as we know, a few documented efforts have been made on cross-lingual hate speech detection in terms of multilingual word embeddings \citep{pamungkas2019cross,pamungkas2020misogyny,arango2020hate}, multilingual multi-aspect hate speech analysis \citep{ousidhoum-etal-2019-multilingual}, and state-of-the-art cross-lingual contextual embeddings like multilingual BERT \citep{devlin2019bert} and XLM-RoBERTa \citep{ranasinghe-zampieri-2020-multilingual,glavas2020xhate}. To further research in broadening the generalisability and suitability of models across languages for hate speech detection \cite{yin2021towards}, we propose a Cross-lingual Capsule Network Learning model with Extra lexical semantics specifically for hate speech (CCNL-Ex), whose two-parallel framework enriches inputs information in both source and target languages. The model can be applied to new languages lacking annotated training data \citep{chen2019multi,pamungkas2019cross} and is able to capture spatial positional relationships between words to improve the generalisability of capsules. It can also be exploited to broaden the detection capacity on linguistically diverse genres such as social media. Our key contributions include: (1) we introduce the first approach to cross-lingual hate speech detection that incorporates capsule networks; (2) we integrate hate-related lexicons into pre-trained word embeddings to investigate their potential to further boost performance; (3) our model yields state-of-the-art performance for all six language pairs under study compared with ten baselines; (4) we perform a comparative study looking into the impact of each layer on our model.

\begin{figure*}[ht]
\centerline{\includegraphics[width=5.7in]{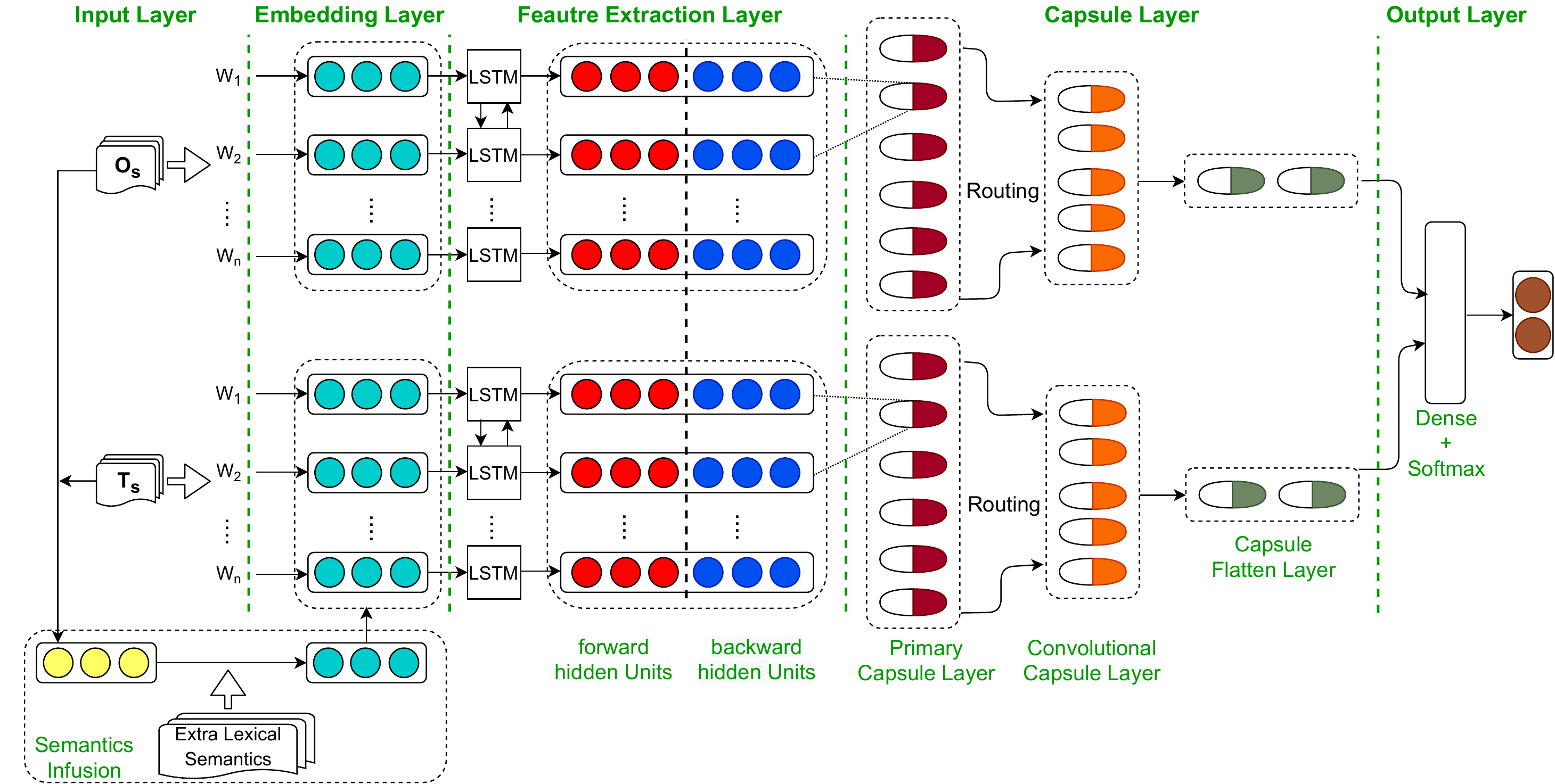}}
\caption{The architecture of CCNL-Ex.}
\label{fig:architecture}
\end{figure*} 

\vspace{-1mm}
\section{Related Work}

\paragraph{\textbf{Cross-lingual Hate Speech Detection.}} With the prevalence of online social media, a range of NLP approaches, especially transfor-mer-based techniques \citep{vaswani2017attention}, have been employed to identify online hate speech \citep{fortuna2018survey,arango2020hate,yin2021towards} or focusing on detecting specific types of hate, such as racism \citep{waseem2016you,davidson2019racial}, sexism \citep{waseem2016hateful,pamungkas2020misogyny}, and cyberbullying \citep{burnap2016us,rezvani2020linking}, but limited to a single language, generally in English. While more mature NLP fields such as sentiment analysis have accumulated substantial efforts by employing cross-lingual learning techniques, work on hate speech detection in cross-lingual scenario has not been explored as much. \citeauthor{basile2018crotonemilano}~\shortcite{basile2018crotonemilano} use SVM with n-grams to tackle English and Italian in a cross-lingual setting in Evalita 2018 and achieve the 15th/2nd position for English/Italian. \citeauthor{pamungkas2019cross}~\shortcite{pamungkas2019cross} propose a joint-learning cross-lingual model with multilingual HurtLex \cite{bassignana2018hurtlex} and MUSE embeddings \citep{conneau2017word}, which outperforms other models using monolingual embeddings \citep{pamungkas2019cross}. Several multilingual multi-aspect approaches are conducted for hate speech \citep{ousidhoum-etal-2019-multilingual} and cross-lingual contextual word embeddings are applied in offensive language identification from English to other languages \citep{ranasinghe-zampieri-2020-multilingual,glavas2020xhate}. Due to the scarcity of cross-lingual resources in this field, some studies tend to generate parallel corpora directly leveraging machine translation resources such as Google Translate \citep{zhou2016cross,pamungkas2019cross,chen2019emoji,pamungkas2020misogyny}, which has proved the effectiveness of the approach. In our study, we consider \citeauthor{pamungkas2019cross}~\shortcite{pamungkas2019cross}'s cross-lingual joint model as a baseline which also uses machine translation as part of their pipeline.

\vspace{-1mm}
\paragraph{\textbf{Capsule Networks.}} Capsule Network is a clustering-like method proposed by \citeauthor{sabour2017dynamic}~\shortcite{sabour2017dynamic}. They replace scalar-output feature detectors of CNNs with vector-output capsules to learn spatial relationships of entities via dynamic routing, improving representations against CNNs. \citeauthor{hinton2018matrix}~\shortcite{hinton2018matrix} then propose a new EM-based iterative routing, which shows potential in image analysis \citep{sabour2017dynamic} and is soon applied to NLP research \citep{zhao2018investigating}. Its use on hate speech detection is however limited \citep{srivastava2018identifying,srivastava2019detecting}. \citeauthor{srivastava2018identifying}~\shortcite{srivastava2018identifying} put forward a capsule-based architecture for aggressive language classification, and further incorporate multi-dimensional capsules for the same task \citep{srivastava2019detecting}. Capsule network has not been considered in cross-lingual hate speech detection so far. Hence, we contribute to gaps in both lines of research bringing together cross-lingual hate speech detection and capsule network.

\vspace{-1mm}
\section{Our Proposed Method: CCNL-Ex}

\subsection{Model Architecture}

Inspired by Capsule Networks built by \citeauthor{sabour2017dynamic}~\shortcite{sabour2017dynamic}, we propose a cross-lingual capsule network learning model with multilingual word embeddings integrated lexical semantics (CCNL-Ex). It is composed of six layers (see Figure \ref{fig:architecture}):%

\vspace{-1mm}
\paragraph{\textbf{Input Layer.}} CCNL-Ex has two parallel capsule-based architectures for bilingual input training data -- $O_s$ in the source language and the parallel translated $T_s$ in the target language.

\vspace{-1mm}
\paragraph{\textbf{Embedding Layer.}} The input data is the sequence of texts and each text consists of a series of words. The input representation is a weight matrix $X \in \mathbb{R}^{e\times V}$ for $e$-dimensional vector of words and vocabulary size of $V$, fine-tuned by absorbing extra hate-related lexical semantic information (see Section \ref{section:integrate-lex}).

\vspace{-1mm}
\paragraph{\textbf{Feature Extraction Layer.}} In each aligned network, we use a Bidirectional Long Short Term Memory (BiLSTM) network \citep{graves2005framewise} as the feature extractor to get contextual relationships from local features. The output of BiLSTM is $h_t = [h^f_t,h^b_t]\in \mathbb{R}^{(2\times k)}$, combined by forward feature $h^f_t$ and backward feature $h^b_t$ with $k$ units.

\vspace{-1mm}
\paragraph{\textbf{Capsule Layer.}} The capsule layer consists of a primary capsule layer and a convolutional capsule layer. The primary capsule layer extracts instantiation parameters to represent spatial position relationships between features, like local order of words and their semantics \citep{zhao2018investigating}. Suppose $W \in \mathbb{R}^{(2\times k) \times d}$ is a shared matrix, where $d$ is the dimensionality of capsules. For the hidden feature $h_t$, we create each capsule $p_i \in \mathbb{R}^d$:
\vspace{-1mm}
\begin{equation}
\label{eq-caps}
    p_i = g(W^Th_t+b)
\end{equation}
\vspace{-0.5mm}
where $g$ is a non-linear squash function to compress the vector length between 0 and 1:
\vspace{-1mm}
\begin{equation}
\label{eq-squash}
    g(s) = \frac{\left \| s \right \|^{2}}{1+\left \| s \right \|^{2}}\frac{s}{\left \| s \right \|}
\end{equation}

\vspace{-0.5mm}

The convolutional capsule layer is connected to capsules in the primary capsule layer. Some primitive routing algorithms, like max pooling in CNNs, only capture features to show whether it exists in a certain position or not, missing more spatial relationships \citep{zhao2018investigating}. The connection weight is learnt by a dynamic routing, which can reduce the loss to make the capsule network more formative and effective, and attach less significance of unrelated or useless content, such as stop words \citep{srivastava2018identifying}. The process of dynamic routing between the primary capsule $u_i$ and the convolutional capsule $v_j$ is as below: 
\begin{equation}
\label{eq-routing1}
    c_{j|i} = softmax(b_{j|i})
\end{equation}
\begin{equation}
\label{eq-routing2}
    v_j = g(\sum_{i}c_{j|i}u_{j|i})
\end{equation}
\begin{equation}
\label{eq-routing3}
    b_{j|i} = b_{j|i} + u_{j|i}\cdot v_j
\end{equation}
where $b_{j|i}$ denotes the connection weight between capsules. After the routing, all output capsules are flattened.

\paragraph{\textbf{Output Layer.}} The final representations from the two parallel architectures are concatenated, using a softmax function to obtain the label probability.

\subsection{Lexical Semantic Knowledge Infusion}
\label{section:integrate-lex}

We fine-tune pre-trained word embeddings by infusing domain-specific lexical semantic knowledge, aiming to obtain domain-aware word representations and enhance model capacity of identifying hate-related content. More specifically, we firstly retrieve five most relevant semantic words from SenticNet \citep{cambria2010senticnet} for each lexical word, and utilise Fasttext embedding model \citep{grave2018learning} to generate the five most similar words for each out-of-vocabulary (OOV) word. Then we apply the similarity learning method proposed by \citeauthor{faruqui2015retrofitting}~\shortcite{faruqui2015retrofitting} to integrate lexicon-derived semantic information into pre-trained word embeddings by minimising distances between a word and its semantically related words.

\section{Experiments}

We investigate cross-lingual hate speech detection as a binary classification task in three different languages --English (EN), Italian (IT) and Spanish (ES)-- and all six possible language pairs involving them: ES$\rightarrow$EN, EN$\rightarrow$ES, IT$\rightarrow$EN, EN$\rightarrow$IT, ES$\rightarrow$IT, and IT$\rightarrow$ES. 

\subsection{Datasets}

We use gender-based hate speech datasets from the Automatic Misogyny Identification (AMI) tasks held at the Evalita 2018\footnote{\url{https://amievalita2018.wordpress.com/data/}} and IberEval 2018\footnote{\url{https://amiibereval2018.wordpress.com/important-dates/data/}} evaluation campaigns. These datasets provided by AMI@Evalita and AMI@IberEval are extracted from the Twitter platform, and constructed under the same annotation scheme for binary labels: misogynistic and non-misogynistic. AMI@Evalita datasets present texts in English and Italian \citep{fersini2018evalita}, while the AMI@Ib-erEval ones are in Spanish and English \citep{fersini2018ibereval}. We utilise English data only from AMI@Evalita to make data size balanced among three languages, as well as for consistency with previous research \citep{pamungkas2019cross} to enable direct comparison. In view of the well-divided training and test sets for each language, we further randomly select 20\% of the training set as the validation set for model fine-tuning process, and finally utilise the whole training set to evaluate model capacity on test set. More details of datasets can be seen in Table \ref{tab:datasets}. We create parallel corpora for separate datasets by directly using Google Translate\footnote{\url{https://translate.google.co.uk/}} to translate all data between source and target languages.

\begin{table}[]
\centering
\caption{Distribution of train, validation and test sets, misogynistic text rate (MTR) in source training and test sets, data sources for three languages.}
\label{tab:datasets}
\begin{tabular}{cccc}
\hline
\textbf{Language}   & \textbf{English (EN)} & \textbf{Spanish (ES)} & \textbf{Italian (IT)} \\ \hline
\textbf{Train}      & 3200                  & 2646                  & 3200                  \\ 
\textbf{Validation} & 800                   & 661                   & 800                   \\ 
\textbf{Test}       & 1000                  & 831                   & 1000                  \\ \hline
\textbf{MTR$_{train}$ (\%)}  & 44.6                  & 49.9                  & 45.7                  \\ 
\textbf{MTR$_{test}$ (\%)}   & 46.0                  & 49.9                  & 50.9                  \\ \hline
\textbf{Source}     & Evalita2018      & IberEval2018      & Evalita2018       \\ \hline
\end{tabular}%
\end{table}

\subsection{Multilingual Lexicons}
\label{ssec:lexicons}

To further assess model performance, we integrate two multilingual domain-related lexicons as extended knowledge into embeddings to explore the possibility of fine-tuning word embeddings and investigate their potential to further boost performance: 

\textbf{HurtLex}\footnote{\url{http://hatespeech.di.unito.it/resources.html}} It is a multilingual hate speech lexicon, containing offensive, aggressive, and hateful words or phrases in over 50 languages and 17 categories. We obtain 6,287 words for English, 3,565 for Spanish and 4,286 for Italian from it.  

\textbf{Multilingual Sentiment Lexicon}\footnote{\url{https://sites.google.com/site/datascienceslab/projects/multilingualsentiment}} Since hate speech often expresses more negative sentiments \citep{mathew2018analyzing}, we utilise a sentiment lexicon, which consists of positive and negative words in 136 languages, and provides 2,955 negative words for English, 2,720 for Spanish and 2,893 for Italian \citep{chen2014building}.

\begin{table*}[htb]
\centering
\caption{Comparison of CCNL and CCNL-Ex over baselines on the six language pairs. The best result in \textbf{bold} and the second best result \underline{underlined}.}
\label{tab:result1}
\begin{tabular}{ccccccc}
\hline
\textbf{Model} & \textbf{ES$\rightarrow$EN} & \textbf{EN$\rightarrow$ES} & \textbf{IT$\rightarrow$EN} & \textbf{EN$\rightarrow$IT} & \textbf{ES$\rightarrow$IT} & \textbf{IT$\rightarrow$ES} \\ \hline
Majority & 0.351 & 0.334 & 0.351 & 0.329 & 0.329 & 0.334 \\ 
SVM & 0.620 & 0.561 & 0.588 & 0.227 & 0.643 & 0.525 \\ 
CNN & 0.598 & 0.613 & 0.592 & 0.275 & 0.636 & 0.607 \\ 
BiLSTM & 0.575 & 0.608 & 0.597 & 0.341 & 0.498 & 0.459 \\ 
CapsNet & 0.616 & 0.559 & 0.601 & 0.323 & 0.555 & 0.611 \\ 
LASER & 0.552 & 0.466 & 0.597 & 0.374 & 0.678 & 0.619 \\
MUSE & 0.592 & 0.491 & 0.618 & 0.400 & 0.717 & \underline{0.666} \\ 
mBERT & 0.567 & 0.580 & 0.568 & 0.399 & 0.648 & 0.618 \\ 
XLM-R & 0.583 & 0.618 & 0.597 & 0.411 & 0.677 & 0.613 \\ 
JL-HL & \underline{0.635} & 0.687 & 0.605 & 0.497 & 0.660 & 0.637 \\ \hline
CCNL & 0.624 & \underline{0.719} & \underline{0.628} & \textbf{0.584} & \textbf{0.735} & \textbf{0.668} \\ 
CCNL-Ex & \textbf{0.651} & \textbf{0.729} & \textbf{0.629} & \underline{0.519} & \underline{0.736} & \textbf{0.670} \\ \hline
\end{tabular}%
\end{table*}

\subsection{Baselines}

We compare both CCNL and CCNL-Ex (with lexicons infused) with ten baselines, including SVM with unigrams features, CNN, BiLSTM and CapsNet with monolingual Fasttext embeddings of translated target data, multilingual embeddings MUSE and LASER fed to a 2-layer feedforward neural network, the state-of-the-art cross-lingual models mBERT and XLM-R covered by a 2-layer feedforward classifier on the output layer, and hate-specific cross-lingual model JL-HL with two inputs proposed by \citeauthor{pamungkas2019cross}~\shortcite{pamungkas2019cross}. All baselines are described as follows:

\vspace{-1.5mm}
\paragraph{\textbf{Majority}} The majority classifier always predicts the most frequent class in the training set. MTR$_{train}$ values for three languages are less than 50\%, which means the majority class is non-misogynistic.
 
\vspace{-1.5mm}
\paragraph{\textbf{SVM}} Support Vector Machine (SVM) aims to determine the best decision boundary between vectors that belong to a given category or not \citep{cortes1995support}; 
\vspace{-1.5mm}
\paragraph{\textbf{CNN}} Consisting of one convolutional layer and one max pooling layer to capture local textual features \citep{kim2014convolutional}; 

\vspace{-1.5mm}
\paragraph{\textbf{BiLSTM}} Composed of forward/backward recurrent neural networks to extract long-term dependencies of a text \citep{cho2014learning}; 

\vspace{-1.5mm}
\paragraph{\textbf{CapsNet}} A single capsule network \citep{sabour2017dynamic} using a convolutional layer to extract n-gram features; 

\vspace{-1.5mm}
\paragraph{\textbf{LASER}} Language-Agnostic SEntence Representations (LASER) aims to calculate and use joint multilingual sentence embeddings across 93 languages \citep{artetxe2019massively};

\vspace{-1.5mm}
\paragraph{\textbf{MUSE}} Multilingual Unsupervised and Supervised Embeddings (MUSE) builds bilingual dictionaries and aligns monolingual word embedding spaces without supervision \citep{conneau2017word}; 

\vspace{-1.5mm}
\paragraph{\textbf{mBERT}} Multilingual BERT\footnote{\url{https://github.com/google-research/bert/blob/master/multilingual.md}} (mBERT) is a variant of BERT \citep{devlin2019bert} that was trained on 104 languages of Wikipedia;

\vspace{-1.5mm}
\paragraph{\textbf{XLM-R}} XLM-RoBERTa is a scaled cross-lingual sentence encoder across 100 languages from Common Crawl \citep{conneau2019unsupervised};

\vspace{-1.5mm}
\paragraph{\textbf{JL-HL}} A joint-learning cross-lingual model proposed by \citeauthor{pamungkas2019cross}~\shortcite{pamungkas2019cross}, a hybrid approach with LSTM architectures which concatenates multilingual lexical features. The source data and translated target data are fed to two parallels separately.

\subsection{Experiment Settings}

For training our model, we use FastText embeddings of dimension 300 trained on the Common Crawl and Wikipedia \citep{grave2018learning}. We use 128 units for forward and backward LSTMs (256 units in total) and 50 units in hidden layer for the feedforward classifier. For capsule networks, we use 10 capsules of dimension 16 and the number of dynamic routing is 5. We use Adam optimiser with 0.0001 learning rate, and set 0.4 for dropout value and 8 for batch size. The model is coded in Keras 2.2.4 and Tensorflow 1.14. We run experiments on the HPC resources of our university, each experiment taking less than one hour. Macro-averaged F1 score is reported as the evaluation metric for all experiments.

\begin{table*}[hbt]
\centering
\caption{CCNL comparative experiment results for six language pairs. The best result in \textbf{bold}.}
\label{tab:result3}
\begin{tabular}{cllllll}
\hline
\textbf{Model} & \multicolumn{1}{c}{\textbf{ES$\rightarrow$EN}} & \multicolumn{1}{c}{\textbf{EN$\rightarrow$ES}} & \multicolumn{1}{c}{\textbf{IT$\rightarrow$EN}} & \multicolumn{1}{c}{\textbf{EN$\rightarrow$IT}} & \multicolumn{1}{c}{\textbf{ES$\rightarrow$IT}} & \multicolumn{1}{c}{\textbf{IT$\rightarrow$ES}} \\ \hline
Results for ablation experiments &  &  &  &  &  & \\ \hline
CCNL-non-parallel & \multicolumn{1}{c}{0.522} & \multicolumn{1}{c}{0.558} & \multicolumn{1}{c}{0.570} & \multicolumn{1}{c}{0.513} & \multicolumn{1}{c}{0.626} & \multicolumn{1}{c}{0.624} \\
CCNL-non-LSTM & \multicolumn{1}{c}{0.373} & \multicolumn{1}{c}{0.609} &  \multicolumn{1}{c}{0.565} & \multicolumn{1}{c}{0.406} & \multicolumn{1}{c}{0.685} & \multicolumn{1}{c}{0.623} \\ 
CCNL-non-Caps & \multicolumn{1}{c}{0.597} & \multicolumn{1}{c}{0.678} &  \multicolumn{1}{c}{0.613} & \multicolumn{1}{c}{0.439} & \multicolumn{1}{c}{0.643} & \multicolumn{1}{c}{0.622} \\ 
CCNL & \multicolumn{1}{c}{\textbf{0.624}} & \multicolumn{1}{c}{\textbf{0.719}} & \multicolumn{1}{c}{\textbf{0.628}} & \multicolumn{1}{c}{\textbf{0.584}} & \multicolumn{1}{c}{\textbf{0.737}} & \multicolumn{1}{c}{\textbf{0.668}} \\ \hline
Results for feature layers &  &  &  &  &  & \\ \hline
CCNL-non-FE & \multicolumn{1}{c}{0.373} & \multicolumn{1}{c}{0.609} &  \multicolumn{1}{c}{0.565} & \multicolumn{1}{c}{0.406} & \multicolumn{1}{c}{0.685} & \multicolumn{1}{c}{0.623} \\ 
CCNL-CNN & \multicolumn{1}{c}{0.521} & \multicolumn{1}{c}{0.592} &  \multicolumn{1}{c}{0.577} & \multicolumn{1}{c}{0.439} & \multicolumn{1}{c}{0.633} & \multicolumn{1}{c}{0.622} \\ 
CCNL-GRU & \multicolumn{1}{c}{0.458} & \multicolumn{1}{c}{\textbf{0.722}} &  \multicolumn{1}{c}{0.613} & \multicolumn{1}{c}{0.411} & \multicolumn{1}{c}{0.715} & \multicolumn{1}{c}{\textbf{0.671}} \\ 
CCNL & \multicolumn{1}{c}{\textbf{0.624}} & \multicolumn{1}{c}{0.719} & \multicolumn{1}{c}{\textbf{0.628}} & \multicolumn{1}{c}{\textbf{0.584}} & \multicolumn{1}{c}{\textbf{0.737}} & \multicolumn{1}{c}{0.668} \\ \hline
\end{tabular}%
\end{table*}

\begin{table*}[htb]
\small
\centering
\caption{Examples for error analysis. Translated texts are presented for non-English instances. Ground truth (GT) and prediction (P) labels are noted -- hateful (1) and non-hateful (0), along with corresponding error types (ET).}
\label{tab:error}
\begin{tabular}{llll}
\hline
\textbf{Text}  & \textbf{GT} & \textbf{P} & \textbf{ET} \\ \hline
\begin{tabular}[c]{@{}l@{}}Analicemos esto: ¿Si te pones unos shorts así, en la calle, ¿qué esperas que te digan? ¿Acoso? ¿O Provocación... \\ \underline{Translation:} Let's analyse this: If you wear shorts like this, in the street, what do you expect them to say? \\Bullying? Or Provocation ...\end{tabular}                   & 1           & 0          & a           \\ \hline
\begin{tabular}[c]{@{}l@{}}tranquille ragazze, tranquilli gay, il Butturini c'ha una morosa che un pezzo di figa mostruosa! \#TVOI\\ \underline{Translation:} quiet girls, quiet gays, Butturini has a girlfriend who is a piece of monstrous pussy! \#TVOI\end{tabular}                                                 & 0           & 1          & b           \\ \hline
\begin{tabular}[c]{@{}l@{}}@user ben sasse is 100\% correct. since 1973, all ive ever heard every two years for elections are hysterical \\women (all a leftist act) about back-alley abortions. this shit is getting old! i didn't hear one other protest \\issue being yelled about i\end{tabular} & 1           & 0          & c           \\ \hline
\begin{tabular}[c]{@{}l@{}}@user ma se la \#culona \#tedesca che predica \#austerit mi sono perso qualcosa\\ \underline{Translation:} @user but if the \#culona \#german preaching \#austerit I missed something\end{tabular}                                                                              & 1           & 0          & d           \\ \hline
\end{tabular}%
\end{table*}

\section{Results}
\label{sec:results}

\subsection{Model Performance}

Results are shown in Table \ref{tab:result1}. CCNL and CCNL-Ex differ in that the latter incorporates lexical semantic features. We can observe that CCNL yields better performance than all baseline models for five out of six language pairs, with the exception of ES$\rightarrow$EN. CCNL-Ex outperforms all ten baselines for all language pairs. These results substantiate the effectiveness of our model with semantic information, highlighting its generalisation capability across three languages. 

Among the ten baselines, the best is JL-HL, whose performance is still always below that of CCNL-Ex. CCNL achieves absolute improvements ranging 7\%-9\% over JL-HL model for two language pairs involving Italian: EN$\rightarrow$IT, and ES$\rightarrow$IT. In addition, the CCNL model has manifested pronounced improvements in terms of separately identifying two classes compared to the majority baseline. We can also observe that CCNL generally achieves a large margin with respect to other baselines like SVM, CNN and BiLSTM (especially for ES$\rightarrow$EN, EN$\rightarrow$IT and IT$\rightarrow$ES), while the baseline MUSE achieves good results for IT$\rightarrow$EN and IT$\rightarrow$ES. It also highlights the effectiveness of the capsule network as an important component in the cross-lingual model compared with other baselines. Furthermore, we observe that CCNL achieves better performance on all six language pairs when we compare it with CapsNet, LASER, mBERT and XLM-R. Possible reasons are that BiLSTM layers enable the proposed model the capability of extracting contextual information compared to the CNN layer, and CCNL takes the spatial features into consideration by sharing the same weight matrix and learning the positional feature difference in high level via the dynamic routing process.

Compared with CCNL, CCNL-Ex performs better for ES$\rightarrow$EN and EN$\rightarrow$ES, and shows a similar performance in three out of six language pairs, which indicates the effectiveness of integrating semantics based on lexicons. The exception to the trend showing a better performance for lexicon-based methods is for the two language pairs that have Italian as the target, namely EN$\rightarrow$IT and ES$\rightarrow$IT, where the base CCNL model with no lexicons performs best. This is likely due to limitations in the Italian language lexicons, and hence reinforces the need to secure high quality lexicons if they are to be incorporated.

\vspace{-1mm}
\subsection{Comparative Experiments}

In order to explore the effect of diverse components in our cross-lingual capsule model on six language-pair tasks, we further implement experiments to assess ablated models compared to our basic framework CCNL and the impact of varying specific components in the feature extraction layer.

\vspace{-1mm}
\subsubsection{\textbf{Framework Ablation Analysis}}

We perform an ablation study for CCNL by dropping one of the two parallel architectures (CCNL-non-parallel), removing the LSTM layer (CCNL-non-LSTM) and removing the Capsule Network layer (CCNL-non-Caps). As shown in Table \ref{tab:result3}, CCNL outperforms all ablated models, demonstrating the combined benefits of all CCNL components. CCNL noticeably outperforms CCNL-non-parallel on all language pairs, highlighting the importance of the two-parallel framework for extracting local features from both source and target texts. Additionally, we can validate the ability of the BiLSTM network to extract contextual information effectively compared with CCNL-non-LSTM, which highlights the effectiveness of the capsule network compared with CCNL-non-Caps.

\subsubsection{\textbf{Impact of Feature Extraction Layer}}

We aim to validate the ability of the BiLSTM network to extract contextual information effectively. We test different feature extraction layers by keeping other components of the CCNL architecture unchanged. We test four flavours of CCNL: CCNL (with LSTM feature extraction layer), CCNL-non-FE (without feature extraction layer), CCNL-CNN (CNN feature extraction instead) and CCNL-GRU (bidirectional GRU feature extraction instead). Results in Table \ref{tab:result3} show that CCNL with the feature extraction layer performs consistently better than those without it, highlighting the importance of extracting local features from the text. Additionally, CCNL also shows improved performance on all tasks when compared with CCNL-CNN, since contextual information plays a significant role in detecting hate speech. CCNL noticeably outperforms CCNL-GRU on four out of six language pairs and achieves similar performance (differences below 0.5\%) on the other two language pairs (EN$\rightarrow$ES and IT$\rightarrow$ES). This is likely due to structural similarities of GRU and LSTM, with the additional complexity of LSTM allowing to capture more informative features in some language pairs.

\vspace{-1.5mm}
\subsection{Error Analysis}

We inspect frequent errors across misclassifications from the test set by the CCNL-Ex model (see Table \ref{tab:error} for examples). We summarise the following four main types of errors: 

\vspace{-1.5mm}
\paragraph{\textbf{(a) Implicit hate: }}Those lacking explicit hateful content or context in the post;
\vspace{-1.5mm}
\paragraph{\textbf{(b) Overuse of hateful words: }}Hateful words can be overused, leading to the over-dependence of the model on these words, while hate targets in posts are confounding and hard to be identified;
\vspace{-1.5mm}
\paragraph{\textbf{(c) Lack of prior information: }}The model cannot identify those contents referring to hate-related event, people or words/phrases with special meanings as it does not possess prior knowledge;
\vspace{-1.5mm}
\paragraph{\textbf{(d) Erroneous translation: }}The use of machine translation can lead to translation errors for important words. Some words used in hashtags cannot be easily translated, which might be regarded as out-of-vocabulary words by the model. 
    
\vspace{-1mm}
\section{Conclusion and Future Work}

We propose a Cross-lingual Capsule Network Learning model integrating Extra hate-related semantic features (CCNL-Ex) for hate speech detection. CCNL, main framework of our model, is composed of two parallel architectures for source and target languages, using BiLSTM to extract contextual features and Capsule Network to capture hierarchically positional relationships. Our model finally leads to state-of-the-art performance for all six language pairs compared with ten competitive baselines. Results show the potential of learning contextual information and spatial relationships of hate speech texts. Given that using machine translation resources such as Google translate is not always a perfect option with very informal language such as those used on social media for hate speech, we will explore approaches to consider culturally grounded context for parallel corpora in future work. Expansive implementations for more languages and datasets are also desired to analyse the generalisability of our model.

\begin{acks}
Aiqi Jiang is funded by China Scholarship Council (CSC). This research utilised Queen Mary's Apocrita HPC facility, supported by QMUL Research-IT. http://doi.org/10.5281/zenodo.438045

\end{acks}

\bibliographystyle{ACM-Reference-Format}
\bibliography{reference}

\end{document}